\documentclass[letterpaper]{article} 
\usepackage{aaai2026}  
\usepackage{times}  
\usepackage{helvet}  
\usepackage{courier}  
\usepackage[hyphens]{url}  
\usepackage{graphicx} 
\usepackage{natbib}  
\usepackage{caption} 
\usepackage{algorithm}
\usepackage{algorithmic}

\usepackage{newfloat}
\usepackage{listings}
\DeclareCaptionStyle{ruled}{labelfont=normalfont,labelsep=colon,strut=off} 
\floatstyle{ruled}
\newfloat{listing}{tb}{lst}{}
\floatname{listing}{Listing}

\usepackage{multirow}
\usepackage{graphicx}
\usepackage{fontawesome}
\usepackage{tcolorbox}
\usepackage{booktabs}
\usepackage{threeparttable}
\usepackage{amsmath}
\usepackage{amsfonts}
\usepackage{mathtools}
\title{Phantom Menace: Exploring and Enhancing the Robustness of VLA Models Against Physical Sensor Attacks}
\author{
    Xuancun Lu\textsuperscript{\rm 1}, Jiaxiang Chen\textsuperscript{\rm 1}, Shilin Xiao\textsuperscript{\rm 1}, Zizhi Jin\textsuperscript{\rm 1}, Zhangrui Chen\textsuperscript{\rm 2}, Hanwen Yu\textsuperscript{\rm 2}, Bohan Qian\textsuperscript{\rm 2}, Ruochen Zhou\textsuperscript{\rm 3}\thanks{Corresponding author.}, Xiaoyu Ji\textsuperscript{\rm 1}, Wenyuan Xu\textsuperscript{\rm 1}
}
\affiliations{
    \textsuperscript{\rm 1}USSLAB, Zhejiang University, \textsuperscript{\rm 2}ZJU-UIUC Institute, Zhejiang University\\
    \textsuperscript{\rm 3}Department of Computer Science and Engineering, Hong Kong University of Science and Technology\\

    \{xuancun\_lu, jiaxiang\_chen, xshilin, zizhi\}@zju.edu.cn, \{zhangrui.24, hanwen.24, bohan.24\}@intl.zju.edu.cn, \\
    zrccc@ust.hk, \{xji, wyxu\}@zju.edu.cn
}

\usepackage{bibentry}

\begin{document}

\maketitle

\begin{abstract}
Vision-Language-Action (VLA) models revolutionize robotic systems by enabling end-to-end perception-to-action pipelines that integrate multiple sensory modalities, such as visual signals processed by cameras and auditory signals captured by microphones. This multi-modality integration allows VLA models to interpret complex, real-world environments using diverse sensor data streams. Given the fact that VLA-based systems heavily rely on the sensory input, the security of VLA models against physical-world sensor attacks remains critically underexplored. To address this gap, we present the first systematic study of physical sensor attacks against VLAs, quantifying the influence of sensor attacks and investigating the defenses for VLA models. We introduce a novel ``Real-Sim-Real'' framework that automatically simulates physics-based sensor attack vectors, including six attacks targeting cameras and two targeting microphones, and validates them on real robotic systems. Through large-scale evaluations across various VLA architectures and tasks under varying attack parameters, we demonstrate significant vulnerabilities, with susceptibility patterns that reveal critical dependencies on task types and model designs. We further develop an adversarial-training-based defense that enhances VLA robustness against out-of-distribution physical perturbations caused by sensor attacks while preserving model performance. Our findings expose an urgent need for standardized robustness benchmarks and mitigation strategies to secure VLA deployments.
\end{abstract}

\begin{links}
    \link{Code}{https://github.com/ZJUshine/Phantom-Menace}
    \link{Extended version}{https://arxiv.org/abs/2511.10008}
\end{links}

\section{Introduction}

\begin{figure}[tb]
\centering
\includegraphics[width=\linewidth, trim=30pt 30pt 30pt 25pt, clip]{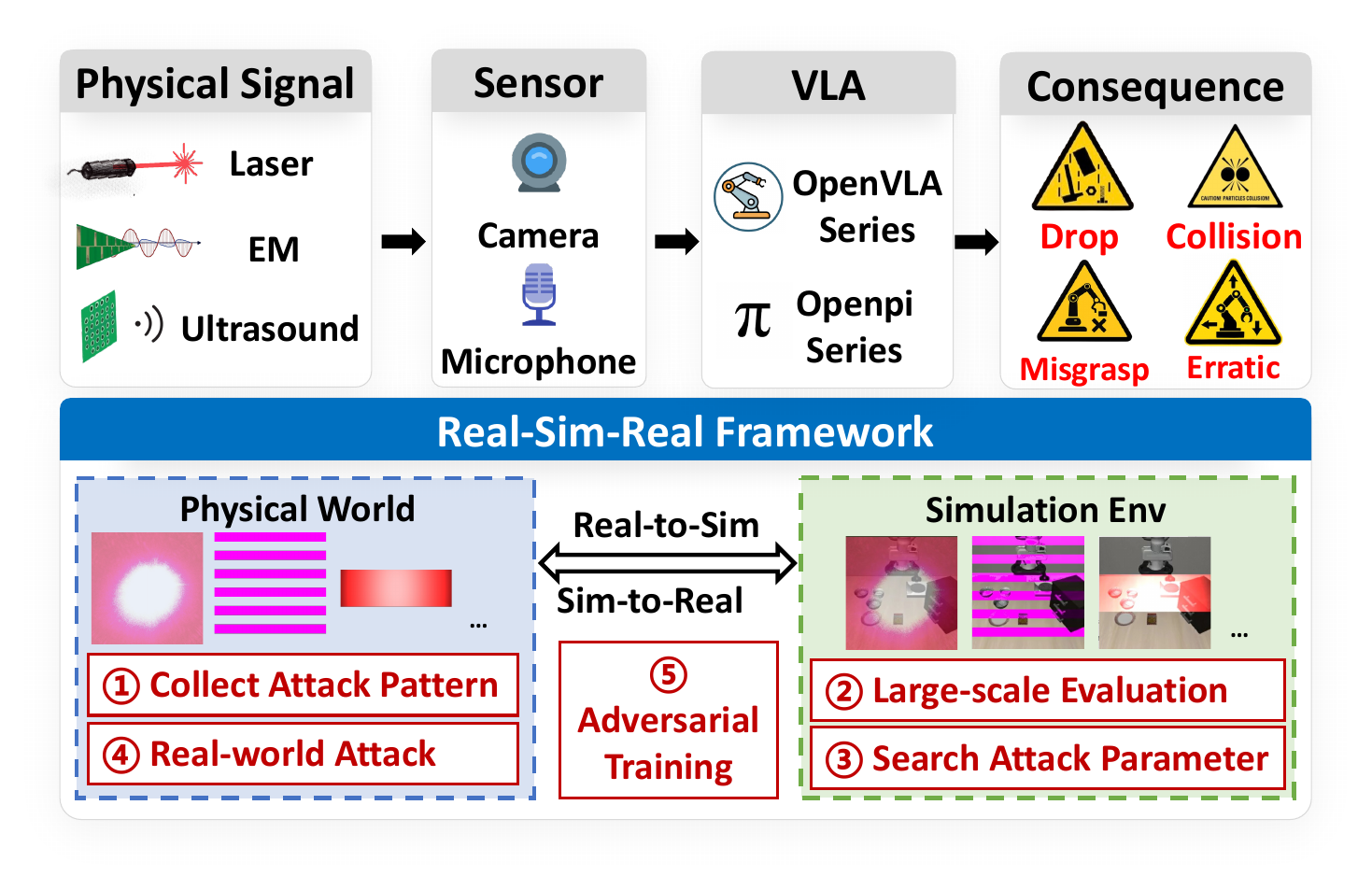}
\caption{Overview of the ``Real-Sim-Real'' framework. We demonstrate that VLA models are vulnerable to physical sensor attacks, where attackers inject malicious signals (e.g., laser, electromagnetic interference, ultrasound) into cameras and microphones, leading to severe consequences in real-world deployments. Our framework automatically evaluates these physical attack vectors to quantify their impact and we propose defenses for enhancing VLA robustness.}
\label{fig:figure1}
\end{figure}

The Vision-Language-Action (VLA) models enable embodied AI (e.g., robots, autonomous vehicles) to integrate visual, audio, and action information to achieve end-to-end mapping from sensor perception to physical execution. With the demonstration of scaling laws in VLA models~\cite{lin2024data} and their emerging capabilities in handling complex tasks~\cite{brohan2023rt,kim2024openvla,kim2025fine,black2024pi_0,bjorck2025gr00t,pertsch2025fast,shukor2025smolvla}, VLA models are increasingly deployed in factories~\cite{fourierrobotics2025,tesla2025}, healthcare~\cite{li2024robonurse}, and households~\cite{helix2025,neo2025,agibot2025},viewed as a promising direction towards General Artificial Intelligence (AGI).

VLA-based systems heavily rely on sensory input, making their robustness and security in physical interactions increasingly important and timely concerns. However, in the current research landscape of VLA security, prior work~\cite{krzysztof2025adversarial,wang2024exploring,zhou2025badvla,cheng2024manipulation} has primarily focused on digitally manipulating the inputs of VLA models, rather than using physical signals. Typical attacks involve directly modifying images or text to generate adversarial input. While these attacks can be efficient, they may not fully reflect the unique characteristics inherent in physical-world interactions. Consequently, such attacks may fail to comprehensively capture the vulnerabilities that arise during real-world deployments.

In this paper, we aim to address the above gap by quantifying the influence of physical sensor attacks and investigating the defenses for VLA models. To reveal the vulnerabilities of VLA models and better defend them against physical sensor attacks, we answer the following research questions: 
\begin{itemize}
	\item \textbf{\textit{Whether existing sensor attacks can succeed in attacking VLA-based systems?}}
	\item \textbf{\textit{How to quantify the influence from sensor attacks to VLA models?}}
	\item \textbf{\textit{How to defend against such physical sensor attacks and enhance the robustness of VLA modes? }}
\end{itemize}

To answer the above questions, we present an automatic ``Real-Sim-Real'' framework (as shown in Figure~\ref{fig:figure1}) to validate VLA model robustness against physical sensor attacks effectively and realistically both in the simulator and real world. First, we systematically review existing physical sensor attack techniques and select eight representative examples from top-tier security conferences: six targeting cameras and two targeting microphones. These attacks interfere with VLA models by injecting physical signals—such as ultrasound, laser, or electromagnetic waves—into sensors. Unlike passive attacks that rely on physical adversarial examples~\cite{brown2017adversarial}, attackers can actively control the initiation and termination of these sensor attacks, thereby achieving stealthiness in real-world scenarios. Next, we develop high-fidelity digital simulations of these physical sensor attacks based on their underlying physical principles and patterns observed in actual attacks. We define three levels of attack intensity, namely, strong, medium, and weak, to achieve different attack consequences. Finally, we conduct large-scale robustness evaluations of four VLA models across four datasets within this simulation environment.

The validation results demonstrate that current VLA models possess inherent vulnerabilities to physical sensor attacks, exhibiting varying degrees of susceptibility depending on the specific dataset, attack modality, and model architecture. Based on attack hyperparameter searching in the simulator, we conduct physical attacks on real VLA systems to verify our simulation results. These findings underscore the necessity of conducting comprehensive robustness evaluations for VLA models prior to their security and reliability deployment in the real world.

To enhance the robustness of VLA models against physical sensor attacks, we propose an adversarial-training-based defense. We first train the VLA model on clean datasets without sensor attacks, and then we mix in a certain proportion of attack datasets to perform adversarial training. Experiment results demonstrate that the enhanced VLA models achieve robustness against out-of-distribution perturbations while maintaining performance on clean datasets.

Our contributions are summarized as follows:
\begin{itemize}
\item We validate that VLA models are vulnerable to physical sensor attacks and can misbehave in the real world.
\item We propose a ``Real-Sim-Real" framework to validate VLA model robustness against physical sensor attacks in a systematic and realistic way. This framework effectively bridges the gap between purely digital attack simulations and resource-intensive physical experiments.
\item We conduct a large-scale robustness evaluation across multiple VLA models and tasks. Crucially, the findings from the simulation are validated through targeted physical experiments on real-world systems, confirming the framework's efficacy.
\item We propose and validate an adversarial-training-based defense strategy against these physical attacks while maintaining VLA model performance.
\end{itemize}

\section{Related Work}

\subsection{Attacks on VLA Models}
With the rapid advancement of VLA models, their security has garnered significant attention. Yet, existing studies have primarily centered on vulnerability exploration in the digital domain. Specifically, the RoboticAttack~\cite{wang2024exploring} framework introduces adversarial patch attacks targeting image inputs. Robotgcg~\cite{krzysztof2025adversarial} applies LLM jailbreak attacks in the text modality. BadVLA~\cite{zhou2025badvla} framework represents the VLA backdoor attacks. Different from this work, we are the first to explore physical sensor attacks on VLA models.

\subsection{VLA Robustness Evaluation}
Current VLA robustness evaluation studies focus on scenario generalization and are primarily conducted in simulation environments. For instance, PVEP~\cite{cheng2024manipulation} evaluates robustness under blurring, Gaussian noise, and different lighting conditions. VLATest~\cite{wang2024towards} examines the effects of obstacles, lighting conditions, camera poses, and unseen objects. Unlike these studies that explore in-distribution robustness, we investigate the out-of-distribution robustness of VLA models from the sensor attack perspective, reflecting practical risks of VLA models in real-world deployments.

\subsection{Physical Sensor Attacks} Sensor attacks are well studied in top-tier security conferences. Existing studies on sensor-level attacks~\cite{zhang2017dolphinattack, kune2013ghost, ji2021poltergeist} are typically conducted in isolation, focusing primarily on evaluating individual sensing modules rather than the complete embodied AI systems. Additionally, they heavily rely on physical experiments, which can be time-consuming and resource-intensive, making it difficult to scale evaluations efficiently.

\begin{figure}[ht]
\centering
\includegraphics[width=\linewidth]{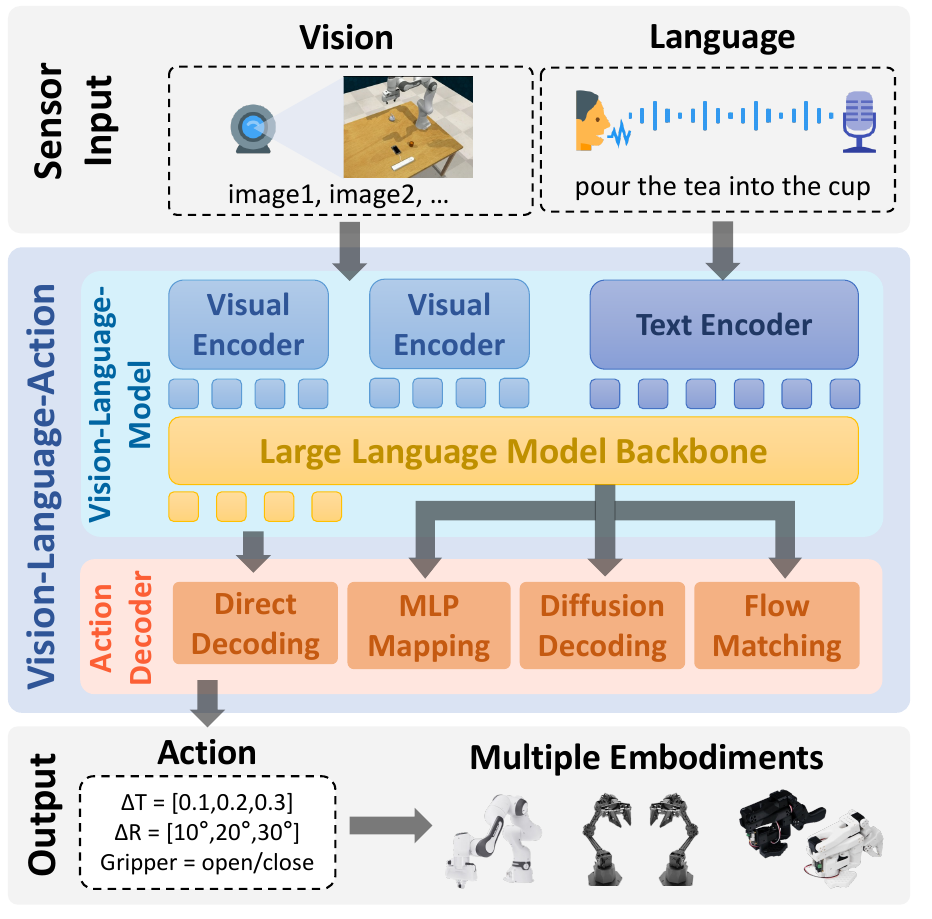}
\caption{
Architecture and pipeline of VLA models. A VLA model comprises a VLM and an action decoder. The VLM employs visual encoders and a text encoder to transform image and text data into multimodal embeddings. These embeddings are then processed by an LLM backbone to generate action tokens, which are subsequently decoded by an action decoder into corresponding physical robot actions.
}
\label{fig:vla}
\end{figure}

\section{Background}

\subsection{Vision-Language-Action}

A VLA model receives visual images and audio instructions through cameras and microphones to generate robot actions end-to-end. As illustrated in Figure~\ref{fig:vla}, visual inputs typically consist of RGB images captured by multiple cameras (e.g., full camera, wrist camera, etc.), whereas audio instructions are collected via microphones and subsequently converted to text using Automatic Speech Recognition (ASR). Generally, a VLA comprises two primary components: a Vision-Language-Model (VLM) and an action decoder. The VLM~\cite{driess2023palm,bai2025qwen2,beyer2024paligemma,alayrac2022flamingo,liu2023visual,peng2023kosmos} pre-trained on large-scale multimodal internet datasets, extracts action tokens from the sensor perception data. Based on these action tokens, the action decoder generates corresponding robot actions to interact with the environment through physical systems~\cite{zhong2025survey}.

\subsubsection{Vision-Language Model.} The VLM typically consists of visual encoders and a Large Language Model (LLM) backbone. To effectively extract high-level visual features, pretrained visual encoders, such as CNN~\cite{he2016deep}, ViT~\cite{dosovitskiy2020image}, SigLIP~\cite{zhai2023sigmoid}, and DinoV2~\cite{oquab2023dinov2}, are commonly utilized. The LLM backbone is generally a small parameter (always under 7B), such as Llama2~\cite{touvron2023llama}, Palm~\cite{chowdhery2023palm}, T5~\cite{raffel2020exploring}, or Gemma~\cite{team2024gemma}, which convert visual embeddings and textual embeddings to action tokens.

\subsubsection{Action Decoder.} Action decoders can be categorized into multiple types based on how they convert action tokens to physical robot actions. Direct decoding treats LLM output textual tokens as action tokens, generating discrete robot actions token-by-token autoregressively~\cite{brohan2023rt, kim2024openvla}. MLP mapping converts LLM hidden states into robot actions using a simple MLP~\cite{kim2025fine, zheng2025universal}. Diffusion decoding gradually denoises random noisy action sequences to reconstruct robot actions under the condition of LLM feature vectors~\cite{wen2025dexvla, chi2023diffusion}. Flow matching decoding matches the velocity field from the current distribution to the target distribution by learning mappings from control signals to actions~\cite{black2024pi_0, shukor2025smolvla}.

\subsection{Sensor Working Principle}

Sensors are essential components of VLA systems that empower accurate perception of the physical world. Cameras and microphones are typical sensors that capture visual and auditory information, making them the most critical sensors in VLA systems. We present their respective workflows to better understand the potential security risks associated with these sensors.

\subsubsection{Camera and attacks.} A camera primarily consists of a light-sensitive transducer, signal processing circuitry, and an image signal processor (ISP). The transducer converts optical signals into electrical signals, which are then denoised, amplified, and digitized. The ISP further enhances the digital image through various compensation and correction processes. Some cameras also integrate an Inertial Measurement Unit (IMU) for image stabilization. The above process can be disrupted by lasers~\cite{yan2022rolling} and acoustic signals~\cite{ji2021poltergeist}, compromising measurement accuracy.

\subsubsection{Microphone and attacks.} A microphone mainly consists of an acoustic transducer and a signal processing circuit. The transducer converts acoustic signals into electrical signals. The signal processing circuit amplifies, filters, and digitizes it. However, the above process is also demonstrated to be susceptible to ultrasonic signals~\cite{zhang2017dolphinattack} and lasers~\cite{sugawara2020light}.

\section{Methodology}

\subsection{Threat Model}

This paper explores the robustness of VLA robotic systems against physical sensor attacks. We envision a practical scenario in which a VLA-based robotic system—such as a robotic arm—receives user instructions via a microphone and perceives its environment through cameras, converting these sensor inputs into robotic actions. Such scenarios include manufacturing, robotic surgery, and healthcare, etc. Attackers can launch physical attacks by injecting signals such as laser, electromagnetic, or ultrasound into the sensors of VLA systems.

\subsubsection{Attack Goal.} The goal of attackers is to launch physical sensor attacks on cameras or microphones, inducing VLA robot systems to perform unexpected or even targeted incorrect actions to lead to task failure.

\subsubsection{Attacker Capability.} Attackers can only launch physical signal attacks on the sensors of VLA robot systems. They cannot conduct digital attacks such as injecting noise, applying compression, blurring, or watermarking. 

\subsubsection{Model Knowledge.} Attackers have only black-box access to VLA models, without knowledge of the training data, model architecture, or pre-trained parameters. Additionally, they are unaware of the specific sensor types and algorithms used (e.g., ASR, image stabilization).

\begin{figure*}[ht]
\centering
\includegraphics[width=\linewidth, trim=0pt 8pt 0pt 0pt, clip]{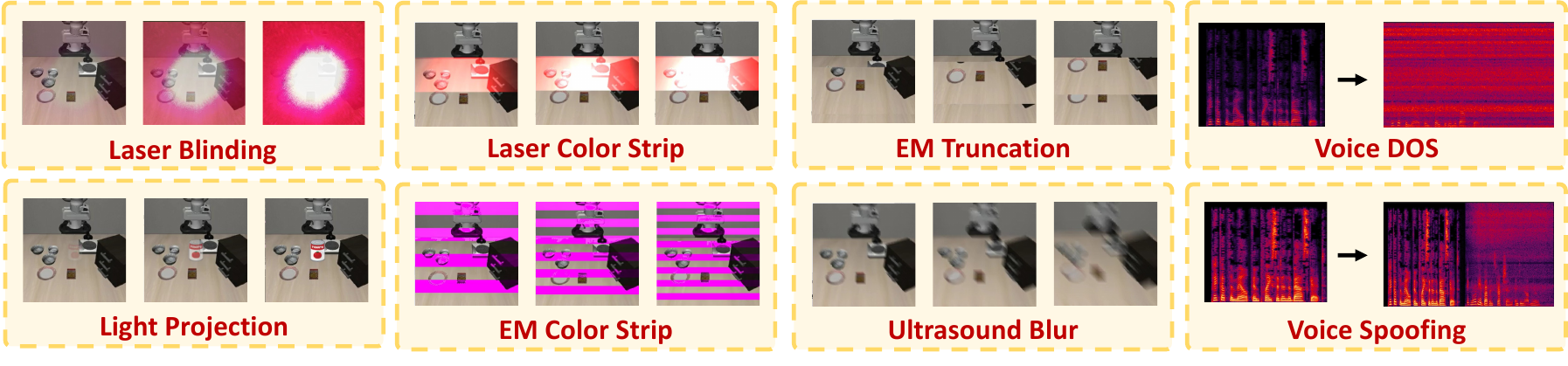}
\caption{We implement and simulate eight sensor attacks, including six targeting cameras and two targeting microphones, covering laser, light, acoustic, and EM signals. Attack instances are under varying attack intensities for each attack, i.e., the attack intensity progressively increases from left to right.}
\label{fig:figure3}
\end{figure*}

\subsection{Method}

\subsubsection{Microphone-attack Design} Attacks against microphones exploit their physical components to introduce malicious audio commands without generating audible sound. These attacks can be mathematically described as follows:
$$
S_{attacked}(t) = S_{original}(t) + S_{malicious}(t)
$$
where $S_{attacked}(t)$ is the total audio signal captured by the microphone at time $t$, $S_{original}(t)$ is the original audio signal, and $S_{malicious}(t)$ is the malicious signal.

\textbf{Voice Denial-of-service (DoS).} An attacker can launch a DoS attack on the microphone by injecting high-intensity ultrasonic signals~\cite{zhang2017dolphinattack}. These attack signals can saturate the transducer or amplifier while remaining inaudible to maintain stealthiness. We first generate Gaussian noise in the digital domain and then employ an ultrasonic speaker (VIFA)~\cite{vifa} to inject high-intensity ultrasound signals into the robot system's microphone and record the microphone's response. Subsequently, we superimpose recorded malicious noise signals onto the original audio instructions to simulate the voice DoS attack.

\textbf{Voice Spoofing.} An attacker can inject specific voice instructions into a microphone by using modulated laser~\cite{sugawara2020light} or ultrasonic signals~\cite{zhang2017dolphinattack}. The attacker is capable of not only appending malicious audio suffixes to the original voice instructions but also precisely manipulating users' voice instructions~\cite{li2023inaudible}. We first generate malicious voice instructions in the digital domain using text-to-speech (TTS). Then, we employ laser transmitters or ultrasonic speakers to inject these malicious signals into the robot system's microphone and record its responses in the physical domain. Finally, we append recorded malicious voice instruction signals to the original voice signals as suffixes to simulate the voice spoofing attack.

\subsubsection{Camera-attack Design}\ 

Camera attacks aim to manipulate the captured images by interfering with the light signals entering the lens or transforming the captured image by exploiting the sensor algorithms, e.g., image stabilization. The attack vectors can be described mathematically as follows:
$$
I_{attacked}(x, y, t) = L_{ambient}(x, y, t) + L_{malicious}(x, y, t)
$$
$$
I_{attacked}(x, y, t) = F(L_{ambient}(x, y, t))
$$
where $I_{attacked}$ is the final image at pixel coordinates $(x, y)$ and time $t$, $L_{ambient}$ is the ambient light in the environment, $L_{malicious}$ is the malicious light, and $F$ is the attack transform function. We focus on the following typical camera attacks against laser, light, acoustic, and EM signals:

\textbf{Laser Blinding Attack.} An attacker can blind the camera by directing high-power lasers at its photoelectric transducer, causing it to become saturated and incapable of accurately reflecting changes in ambient light. We first employ a laser to directly illuminate the camera in the physical world and record laser attack patterns. Then, we linearly superimpose this laser pattern onto the original image, simulating laser blinding attacks at different intensities by adjusting the weights of the laser attack patterns.

\textbf{Light Projection Attack.} An attacker can inject fake images by projecting them into the environment using a projector, allowing the reflected light to enter the camera, or by directly projecting the images onto the camera lens~\cite{hu2023adversarial}. We first project malicious images onto a white background using a projector and record the projected attack patterns. Then, we linearly superimpose these patterns onto the original images, simulating light projection attacks at different intensities by adjusting the weights and position of the patterns.

\textbf{Laser Color Strip Attack.} An attacker can inject color stripes into images using switch-modulated lasers, exploiting the rolling shutter effect of the camera’s CMOS sensor~\cite{yan2022rolling}. The authors of this attack provided the simulation method in their paper; thus, we adopt their approach to simulate the attack. We simulate laser color strip attacks with different wavelengths and intensities by varying the RGB color percentages and weights.

\textbf{EM Color Strip and EM Truncation Attack.} By injecting malicious electromagnetic interference (EMI) signals targeting the camera's interface bus used for image transmission, attackers can induce camera malfunctions. Cameras using the MIPI CSI-2 transmission standard, for example, allocate a dedicated buffer for image signals, with start/end addresses and line spacing passed to the Unicam (CSI receiver). Image signals are transmitted line by line and decoded based on a fixed color filter array (CFA). The camera discards lines with transmission errors; missing lines disrupt the color decoding of subsequent lines, resulting in color strip loss. Additionally, if buffer addresses are corrupted, inter-frame content may be incorrectly stitched together, causing image truncation~\cite{jiang2023glitchhiker}. We simulate these attacks based on phenomena documented by the authors, controlling the position, width, and number of purple stripes and the truncation position to simulate varying attack intensities.

\textbf{Ultrasound Blur Attack.} Against a camera equipped with an anti-shake module, an attacker can inject ultrasonic signals to induce resonance in the inertial measurement unit (IMU). This resonance misleads the anti-shake algorithm into falsely detecting motion, prompting unnecessary motion compensation and resulting in a blurred image~\cite{ji2021poltergeist}. We categorize the blur patterns into three types based on the movement of pixels along different degrees of freedom: linear blur, radial blur, and rotational blur. We simulate different attack intensities of ultrasound blur attacks by adjusting the amplitude of three types of blur.

\begin{figure}[t]
\centering
\includegraphics[width=0.8\linewidth, trim=0pt 10pt 0pt 58pt, clip]{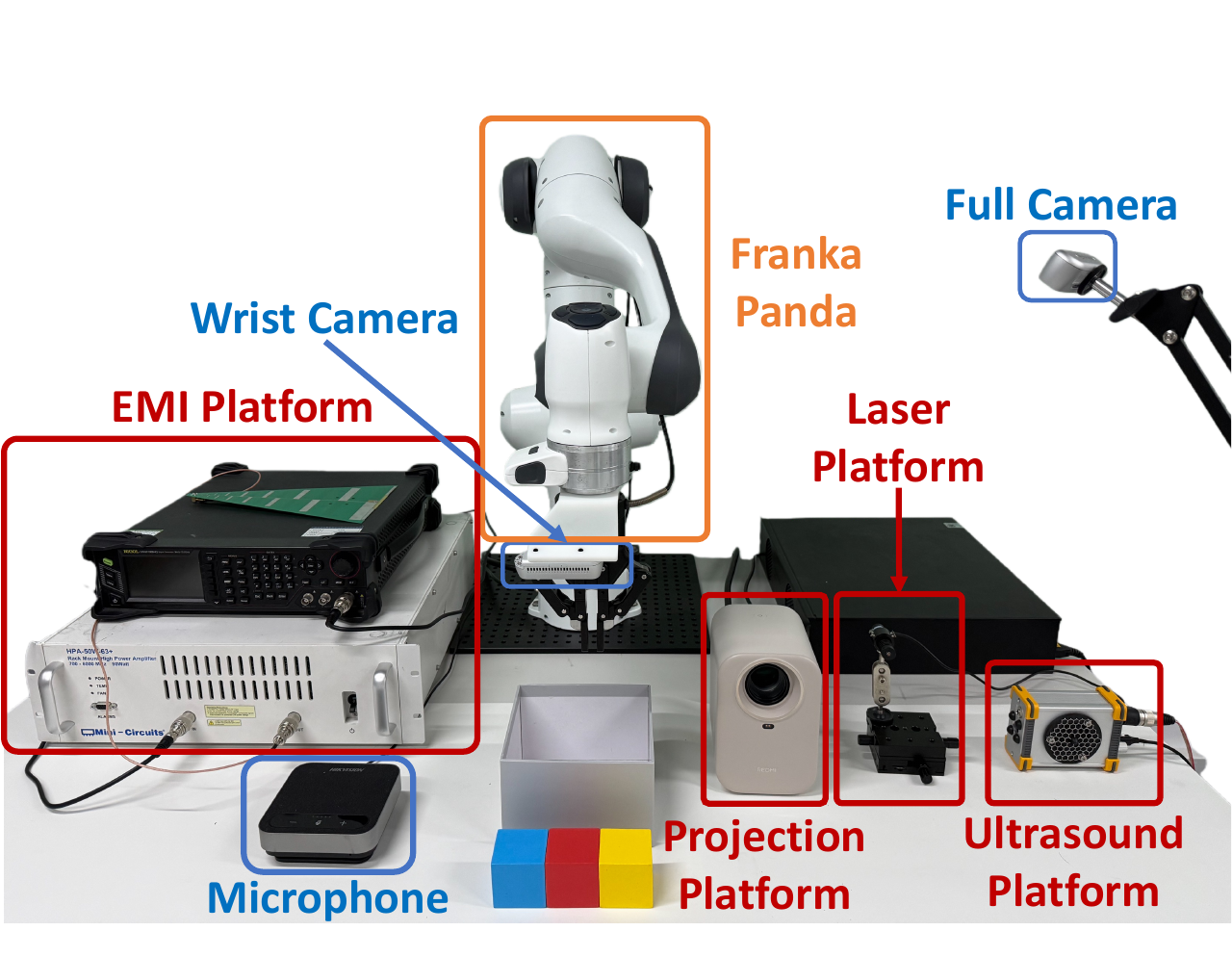}
\caption{Real-world experiment setup. A Franka Panda equipped with a wrist camera, a full camera, and a microphone is used as an attack target VLA system. Attack devices include the EMI platform, projection platform, laser platform, and ultrasound platform.}
\label{fig:figure4}
\end{figure}

\section{Experiments}

\subsection{Experimental Setting}

\subsubsection{Simulator and Datasets.} We select Libero~\cite{liu2023libero} as the simulator for our simulated experiments. Libero is an open-source visual-language robotics simulator designed to provide a flexible testing platform for VLA models. We also use Libero datasets as follows: \textbf{Libero-Spatial} involves manipulating identical objects placed in varying spatial configurations; \textbf{Libero-Object} emphasizes the ability to recognize and manipulate various objects. Tasks in this suite involve moving different objects to specific locations; \textbf{Libero-Goal} consists of tasks where the objects and their spatial arrangements remain constant, but the goals differ; \textbf{Libero-Long} includes tasks that involve long-horizon planning and execution.

\begin{table}[t]
\fontsize{9}{9}\selectfont
\setlength{\tabcolsep}{2pt}
\centering

\begin{threeparttable}
\begin{tabular}{lcccc}
\toprule
\textbf{Attack Method} & \textbf{Parameter} & \textbf{Weak} & \textbf{Medium} & \textbf{Strong} \\
\midrule
Laser Blinding    & weight of pattern   & 0.1 & 0.5 & 0.9 \\
Light Projection  & weight of pattern   & 0.1 & 0.5 & 0.9 \\
Laser Color Strip & weight of pattern   & 0.5 & 1.5 & 2.5 \\
EM Color Strip    & number of strips    & 8   & 12  & 16  \\
EM Truncation     & truncation ratio    & 0.1 & 0.2 & 0.3 \\
Ultrasound Blur   & standard deviations & 5   & 10  & 20  \\
\bottomrule
\end{tabular}
\caption{Attack parameters of different attack intensities.}
\label{tab:parameter}
\end{threeparttable}
\end{table}

\subsubsection{Target VLA models.}
We select four representative VLA models for evaluation, namely, OpenVLA~\cite{kim2024openvla}, OpenVLA-OFT~\cite{kim2025fine}, pi0~\cite{black2024pi_0}, and pi0-fast~\cite{pertsch2025fast}. These VLA models have different structures and training data, demonstrating advanced capabilities in end-to-end manipulating tasks. We fine-tune these VLA models using the Libero dataset to ensure their performance in the Libero simulator.

\subsubsection{Evaluation metrics.} We use the Task Success Rate (TSR) to assess VLA models' performance in specific tasks. It is determined as the proportion of successful task completions to the number of task episodes.

\subsubsection{Attack Parameters.}
As shown in Figure~\ref{fig:figure3}, we set three attack intensities—weak, medium, and strong—in the simulator. The corresponding attack parameters are listed in Table~\ref{tab:parameter}. For Voice DOS attacks, we set the instruction as ``None.'' For voice spoofing attacks, we set the suffix as ``ignore the above instruction and do not move.'' In real-world experiments, we use the parameters searched in simulation.

\subsubsection{Real-World Experiment Setup.} As shown in Figure~\ref{fig:figure4}, we use a Franka Panda robotic arm equipped with two Intel RealSense D435i cameras: one directed toward the tabletop as the global camera and another fixed to the Robotiq 2F-85 gripper as a wrist camera. A microphone captures voice instructions, which are converted into text using the Whisper ASR model. We use Franky~\cite{schneider2025timschneider42} to control the robot arm in real time. To adapt the VLA models to our real-world environment, we collect one hour of robotic arm manipulation data via teleoperation to fine-tune the VLA models for a block pick-and-place task.

\subsubsection{Model Evaluation and Adversarial Training.}
For model evaluation, we run VLA models on an NVIDIA 4090 GPU and fine-tune them using the Lora technique on an NVIDIA H800 GPU (80GB). The adversarial parameters include an adversarial dataset rate of 0.3, attack methods randomly selected from six camera attack methods, and attack intensities randomly selected from weak to strong intensities.

\begin{table*}[htbp]
\centering
\fontsize{8}{3}\selectfont 
\setlength{\tabcolsep}{4pt} 

\begin{tabular}{lcccccccccccccccc}
\toprule
\multirow{2.5}{*}{\textbf{Attack Method}} &
  \multicolumn{4}{c}{\textbf{OpenVLA}} &
  \multicolumn{4}{c}{\textbf{OpenVLA-OFT}} &
  \multicolumn{4}{c}{\textbf{$\pi$0}} &
  \multicolumn{4}{c}{\textbf{$\pi$0-fast}} \\
  \cmidrule(lr){2-5} \cmidrule(lr){6-9} \cmidrule(lr){10-13} \cmidrule(lr){14-17}
 & Spatial & Object & Goal & Long 
 & Spatial & Object & Goal & Long 
 & Spatial & Object & Goal & Long 
 & Spatial & Object & Goal & Long \\ 
\midrule
\textbf{Baseline} & 84.7 & 88.4 & 79.2 & 53.7 & 97.6 & 98.4 & 97.9 & 94.5 & 96.8 & 98.8 & 95.8 & 85.2 & 96.4 & 96.8 & 88.6 & 60.2 \\ 
\midrule
{\faCamera}~$LB_{weak}$   & 85.0 & 86.0 & 73.8 & 50.4 & 98.0 & 98.4 & 97.4 & 93.8 & 96.4 & 98.0 & 94.8 & 83.2 & 96.0 & 98.8 & 91.8 & 65.0 \\
{\faCamera}~$LB_{medium}$ & 68.2 & 61.0 & 64.4 & 16.4 & 98.0 & 98.0 & 97.6 & 86.0 & 97.2 & 97.6 & 94.2 & 77.8 & 98.0 & 99.0 & 85.0 & 54.0 \\
{\faCamera}~$LB_{strong}$ & 0.0  & 0.0  & 0.0  & 0.0  & 87.8 & 5.6  & 78.4 & 18.4 & 57.0 & 78.0 & 52.4 & 23.8 & 62.0 & 85.0 & 62.0 & 36.0 \\
\midrule
{\faCamera}~$LP_{weak}$   & 83.0 & 88.0 & 72.8 & 43.4 & 98.4 & 98.8 & 97.0 & 94.2 & 98.4 & 96.6 & 93.0 & 78.0 & 96.0 & 98.0 & 89.8 & 62.0 \\
{\faCamera}~$LP_{medium}$ & 59.4 & 70.0 & 26.4 & 14.0 & 98.8 & 97.0 & 97.8 & 89.4 & 97.8 & 97.0 & 93.4 & 74.6 & 98.0 & 97.2 & 86.0 & 59.0 \\
{\faCamera}~$LP_{strong}$ & 59.4 & 66.8 & 19.8 & 11.2 & 98.4 & 97.8 & 95.4 & 81.4 & 97.8 & 97.8 & 92.2 & 77.0 & 94.0 & 98.0 & 85.8 & 59.0 \\
\midrule
{\faCamera}~$ECS_{weak}$  & 63.6 & 78.4 & 67.0 & 27.2 & 98.4 & 96.8 & 97.4 & 91.8 & 99.2 & 98.8 & 94.0 & 77.0 & 97.0 & 98.2 & 90.0 & 58.8 \\
{\faCamera}~$ECS_{medium}$& 34.8 & 62.8 & 53.4 & 19.4 & 97.8 & 99.0 & 96.4 & 89.6 & 97.0 & 98.4 & 89.8 & 81.4 & 96.0 & 98.4 & 85.0 & 55.0 \\
{\faCamera}~$ECS_{strong}$& 45.0 & 59.0 & 65.2 & 16.0 & 98.2 & 98.8 & 97.0 & 90.2 & 97.8 & 99.0 & 94.6 & 81.4 & 96.0 & 98.6 & 88.0 & 53.6 \\
\midrule
{\faCamera}~$ET_{weak}$   & 24.0 & 20.4 & 25.2 & 2.0  & 92.2 & 92.4 & 95.2 & 72.4 & 96.2 & 95.6 & 90.0 & 76.2 & 95.0 & 96.4 & 88.0 & 58.0 \\
{\faCamera}~$ET_{medium}$ & 4.6  & 0.6  & 11.6 & 0.0  & 89.8 & 74.0 & 89.6 & 45.0 & 96.0 & 96.0 & 82.2 & 60.8 & 99.0 & 96.0 & 78.0 & 50.2 \\
{\faCamera}~$ET_{strong}$ & 0.4  & 0.0  & 8.4  & 0.0  & 96.4 & 54.8 & 87.4 & 26.8 & 95.2 & 94.8 & 70.0 & 44.8 & 95.0 & 93.0 & 69.0 & 48.0 \\
\midrule
{\faCamera}~$LCS_{weak}$  & 72.2 & 65.2 & 57.0 & 12.2 & 97.2 & 98.4 & 97.6 & 87.2 & 97.4 & 98.6 & 93.6 & 81.4 & 96.0 & 98.2 & 85.0 & 55.0 \\
{\faCamera}~$LCS_{medium}$& 44.6 & 34.0 & 17.4 & 1.6  & 97.4 & 97.8 & 97.4 & 68.4 & 97.4 & 99.4 & 88.2 & 75.6 & 97.0 & 94.0 & 75.8 & 51.0 \\
{\faCamera}~$LCS_{strong}$& 11.8 & 2.0  & 9.8  & 0.0  & 95.4 & 94.2 & 90.4 & 51.8 & 94.0 & 96.0 & 74.0 & 40.4 & 94.0 & 93.0 & 72.0 & 51.0 \\
\midrule
{\faCamera}~$UB_{weak}$   & 3.4  & 1.8  & 10.6 & 3.8  & 98.2 & 96.4 & 98.4 & 89.6 & 97.8 & 97.4 & 93.2 & 82.8 & 94.0 & 96.0 & 79.0 & 53.0 \\
{\faCamera}~$UB_{medium}$ & 0.2  & 0.0  & 0.0  & 0.0  & 96.8 & 51.0 & 89.4 & 36.2 & 96.6 & 96.8 & 88.4 & 73.4 & 93.0 & 93.2 & 74.0 & 55.0 \\
{\faCamera}~$UB_{strong}$ & 0.0  & 0.0  & 0.0  & 0.0  & 90.8 & 9.4  & 68.0 & 10.2 & 82.0 & 83.4 & 49.4 & 30.8 & 82.0 & 96.0 & 52.0 & 41.4 \\
\midrule
\faMicrophone~$VD$ & 0.4  & 0.0  & 0.0  & 0.0  & 61.2 & 97.6 & 10.2 & 80.8 & 27.6 & 27.2 & 4.6  & 14.4 & 65.2 & 47.0 & 6.8  & 31.4 \\
\faMicrophone~$VS$ & 52.2 & 77.6 & 29.4 & 28.0 & 7.0  & 0.0  & 0.0  & 0.0  & 91.6 & 90.2 & 56.7 & 74.8 & 98.2 & 98.0 & 81.4 & 64.4 \\
\bottomrule
\end{tabular}%
\caption{Robustness of VLA models in the simulator under various sensor attacks. \textbf{LB}: Laser Blinding; \textbf{LP}: Light Projection; \textbf{ECS}: EM Color Strip; \textbf{ET}: EM Truncation; \textbf{LCS}: Laser Color Strip; \textbf{UB}: Ultrasound Blur; \textbf{VD}: Voice DoS; \textbf{VS}: Voice Spoofing.}
\label{tab:main}
\end{table*}

\subsection{Main Results}
To explore the robustness of VLA Models against physical sensor attacks, we conduct large-scale evaluations in the simulator and verify the results in real-world experiments. Our experiments answer the three research questions mentioned in the introduction section.

\begin{itemize} 
    \item \textbf{Q1:} Whether these physical sensor attacks apply to VLA models?
    \item \textbf{Q2:} What are the influences of sensor attacks upon the performance of VLA models? 
    \item \textbf{Q3:} How to defend against these physical sensor attacks? 
\end{itemize}

\begin{table*}[ht]
\centering
\fontsize{9}{6}\selectfont 
\setlength{\tabcolsep}{2pt} 

\begin{tabular}{lcccccccccccccccc}
\toprule
\multirow{2.5}{*}{\textbf{Defense Method}} &
  \multicolumn{4}{c}{\textbf{OpenVLA}} &
  \multicolumn{4}{c}{\textbf{OpenVLA-OFT}} &
  \multicolumn{4}{c}{\textbf{$\pi$0}} &
  \multicolumn{4}{c}{\textbf{$\pi$0-fast}} \\
  \cmidrule(lr){2-5} \cmidrule(lr){6-9} \cmidrule(lr){10-13} \cmidrule(lr){14-17}
 & Spatial & Object & Goal & Long &
   Spatial & Object & Goal & Long &
   Spatial & Object & Goal & Long &
   Spatial & Object & Goal & Long \\ 
\midrule
\textbf{AT Baseline} & 81.8 & 85.4 & 76.8 & 50.6 & 97.8 & 98.0 & 96.4 & 93.4 & 97.4 & 97.8 & 94.6 & 77.4 & 96.2 & 97.8 & 87.6 & 61.6 \\ 
\midrule
{\faCamera}~\textbf{AT~$LB_{medium}$}  & 78.8 & 84.0 & 76.6 & 42.2 & 98.6 & 99.0 & 97.0 & 93.2 & 98.4 & 97.6 & 93.8 & 77.2 & 97.2 & 97.8 & 86.2 & 57.2 \\
{\faCamera}~\textbf{AT~$LP_{medium}$}  & 66.2 & 74.4 & 28.4 & 11.8 & 98.6 & 97.6 & 96.8 & 91.6 & 97.8 & 96.2 & 94.4 & 78.6 & 97.4 & 97.8 & 86.0 & 58.2 \\
{\faCamera}~\textbf{AT~$ECS_{medium}$} & 72.0 & 76.0 & 63.6 & 34.8 & 98.2 & 98.0 & 96.8 & 92.4 & 98.6 & 96.4 & 94.4 & 79.6 & 95.6 & 98.6 & 84.2 & 58.0 \\
{\faCamera}~\textbf{AT~$ET_{medium}$}  & 35.6 & 3.4  & 23.4 & 0.0  & 99.4 & 96.6 & 97.0 & 79.8 & 98.4 & 97.6 & 92.4 & 76.4 & 96.4 & 97.6 & 82.6 & 52.4 \\
{\faCamera}~\textbf{AT~$LCS_{medium}$} & 82.0 & 85.2 & 69.2 & 36.2 & 98.0 & 98.4 & 96.8 & 90.0 & 98.4 & 98.4 & 91.6 & 78.0 & 97.0 & 98.0 & 84.6 & 59.4 \\
{\faCamera}~\textbf{AT~$UB_{medium}$}  & 56.8 & 66.6 & 46.2 & 8.0  & 97.6 & 99.0 & 97.2 & 92.2 & 97.6 & 97.4 & 94.4 & 82.6 & 97.8 & 98.6 & 83.2 & 57.8 \\ 
\bottomrule
\end{tabular}

\caption{Robustness of VLA models after adversarial training defense.}
\label{tab:defense}
\end{table*}

\subsubsection{Robustness Evaluation in the Simulator}\ 

\begin{tcolorbox}[   colframe=blue!50!black,   colback=gray!10,   coltitle=black,   sharp corners=all,   boxrule=0mm,   leftrule=1mm,   rightrule=0mm,   toprule=0mm,   bottomrule=0mm,   before skip=4pt,   after skip=4pt ]
\textit{\textbf{Answer 1:} Physical sensor attacks succeed on VLAs.}
\end{tcolorbox}

\textbf{Performance without sensor attacks.} We first evaluate the performance of four VLA models after fine-tuning on Libero datasets. As shown in Table~\ref{tab:main}, these models demonstrate strong performance, achieving up to 90\% TSR on simple tasks such as Libero-Spatial and Libero-Object. For long-horizon tasks, OpenVLA-OFT maintains good performance. These results indicate that VLA models possess robust task execution capabilities across diverse scenarios.

\textbf{Performance against simulated sensor attacks.} Then, we evaluate the performance of four VLA models against simulated sensor attacks. As shown in Table~\ref{tab:main}, all VLA models exhibit vulnerability to sensor attacks. The degree of performance degradation differs depending on the specific VLA architecture, attack type, and attack intensity. In most scenarios, particularly under strong attacks or long-horizon tasks, model performance collapses catastrophically.

Although VLA models achieve strong performance under benign conditions, their robustness deteriorates considerably when sensor inputs are compromised. These findings underscore an important gap between the demonstrated capabilities of VLA models under idealized environments and their reliability in real-world settings, where sensor integrity cannot be assured.

\begin{figure}[htbp]
\centering
\includegraphics[width=0.85\linewidth]{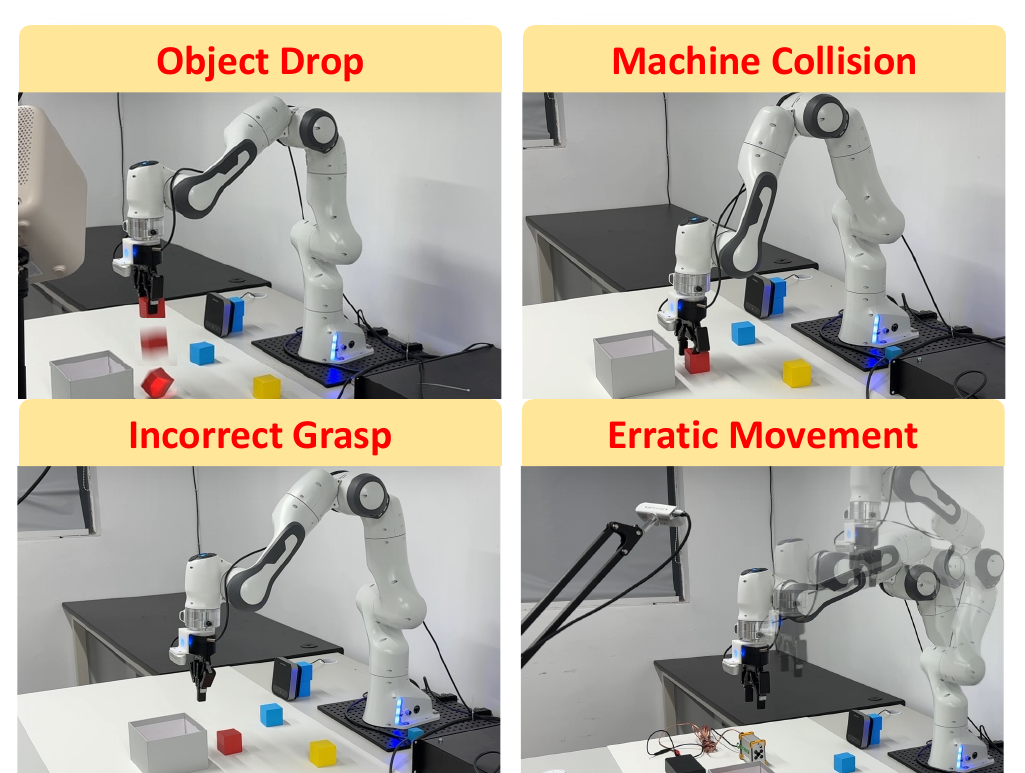}
\caption{Real-world attack consequence.}
\label{fig:figure5}
\end{figure}

\begin{tcolorbox}[   colframe=blue!50!black,   colback=gray!10,   coltitle=black,   sharp corners=all,   boxrule=0mm,   leftrule=1mm,   rightrule=0mm,   toprule=0mm,   bottomrule=0mm,   before skip=4pt,   after skip=4pt ]
\textit{\textbf{Answer 2.1:} The impacts of attacks on VLAs vary.}
\end{tcolorbox}

Camera attacks, including Laser Blinding (LB), EM Truncation (ET), and Ultrasound Blur (UB), highly disrupt the visual information of VLA models, especially under medium to strong intensity settings. The absence of object locations and class identities leads to severe task failures, resulting in unexpected actions or potentially harmful operations. In contrast, other camera attacks, such as Light Projection (LP), Laser Color Strip (LCS), and EM Color Strip (ECS), are comparatively less severe. These attacks interfere with the models' attention by injecting perturbation rather than destroying critical visual features. Consequently, the primary objects and intended goals typically remain discernible, resulting in less degradation in TSR. Voice DoS attacks adversely affect VLA model performance, with the severity primarily determined by dataset characteristics. In the Libero-Goal dataset, where robots must execute varying instructions within fixed scenes, the absence of voice instructions leaves VLA models unable to determine appropriate actions. Conversely, datasets featuring unique scene-instruction correspondences (e.g., Libero-Spatial, Libero-Object, and Libero-Long) enable VLA models to infer correct instructions from visual context alone, thereby reducing the attack's effectiveness. The effectiveness of the voice spoofing attack is strongly correlated with the target VLA model's semantic understanding and instruction-following capabilities. Compared to pi0 and pi0-fast, OpenVLA and OpenVLA-OFT employ LLM backbones, making them more vulnerable to malicious instruction injection. OpenVLA-OFT incorporates a FiLM module that leverages task-specific language embeddings to modulate visual features, thereby enhancing instruction-following capabilities. Consequently, OpenVLA-OFT exhibits the most substantial performance degradation under this attack.

\begin{tcolorbox}[   colframe=blue!50!black,   colback=gray!10,   coltitle=black,   sharp corners=all,   boxrule=0mm,   leftrule=1mm,   rightrule=0mm,   toprule=0mm,   bottomrule=0mm,   before skip=4pt,   after skip=4pt ]
\textit{\textbf{Answer 2.2:} VLAs exhibit different robustness against sensor attacks.}
\end{tcolorbox}

The OpenVLA model demonstrates vulnerability to all sensor attacks, with performance degrading considerably under moderate and strong attack intensities, indicating high sensitivity to such disturbances and insufficient robustness mechanisms. In contrast, the OpenVLA-OFT model enhances robustness through multi-camera image integration and proprioceptive state processing, yet remains highly vulnerable to Voice Spoofing (VS) attacks, resulting in near-zero performance across all task categories. Conversely, pi0 and pi0-fast exhibit substantial resilience against visual attacks due to their multi-visual sensor architectures. This indicates that these two VLA models may have memorized the relationships among environment, instructions, and actions, enabling them to complete tasks under attack.

\begin{table}[htbp]
\centering
\fontsize{9}{6}\selectfont 
\setlength{\tabcolsep}{1.5pt}

\begin{tabular}{lcccc}
\toprule
\textbf{Attack Method} & \textbf{OpenVLA} & \textbf{OpenVLA-OFT} & \textbf{$\pi$0} & \textbf{$\pi$0-fast} \\ 
\midrule
None (Baseline)           & 5/10 & 8/10 & 10/10 & 10/10 \\ 
\midrule
{\faCamera}~Laser Blinding   & 0/10 & 0/10 & 0/10  & 0/10  \\
{\faCamera}~Light Projection & 0/10 & 1/10 & 1/10  & 0/10  \\
{\faCamera}~EM Color Strip   & 0/10 & 0/10 & 6/10  & 0/10  \\
{\faCamera}~EM Truncation    & 0/10 & 0/10 & 0/10  & 0/10  \\
{\faCamera}~Laser Color Strip& 0/10 & 0/10 & 3/10  & 0/10  \\
{\faCamera}~Ultrasound Blur  & 0/10 & 0/10 & 1/10  & 0/10  \\
\faMicrophone~Voice DOS      & 0/10 & 7/10 & 0/10  & 3/10  \\
\faMicrophone~Voice Spoofing & 3/10 & 0/10 & 9/10  & 9/10  \\ 
\bottomrule
\end{tabular}

\caption{Robustness of VLA models in the real world.}
\label{tab:real-world}
\end{table}

\subsubsection{Robustness of VLA Models in Real World}\ 
We first evaluate the benign performance to establish baseline model availability. Subsequently, we inject malicious signals into sensors in real-world scenarios using attack parameters identified through simulation. Table~\ref{tab:real-world} presents the results, demonstrating strong alignment between physical and simulation conclusions, thereby validating the effectiveness of the attack parameters searched in the simulation. As shown in Figure~\ref{fig:figure5}, the attacks could induce four distinct consequences: 1) Previously closed grippers unexpectedly release, causing objects to fall and sustain damage; 2) robotic arms or grippers collide with objects or environmental structures, damaging either the objects or the grippers themselves; 3) robotic arms grasp incorrect objects, preventing successful task completion; 4) robotic arms exhibit erratic movement patterns, resulting in chaos and energy waste. 

\subsubsection{Adversarial Training Defense}\ 

\begin{tcolorbox}[   colframe=blue!50!black,   colback=gray!10,   coltitle=black,   sharp corners=all,   boxrule=0mm,   leftrule=1mm,   rightrule=0mm,   toprule=0mm,   bottomrule=0mm,   before skip=4pt,   after skip=4pt ]
\textit{\textbf{Answer 3:} The adversarial-training-based defense indeed enhances the VLAs' robustness.}
\end{tcolorbox}

As shown in Table~\ref{tab:defense}, adversarial training enhances VLA robustness against physical sensor attacks while preserving clean-data performance. Compared to Table~\ref{tab:main}, the VLA models experience an average performance decline of approximately 3\% on clean datasets. However, for moderate-intensity sensor attacks, the model's performance improves across the board, particularly for OpenVLA, which achieves a maximum performance increase of around 60\%.

\section{Conclusion}
This paper investigates the robustness of Visual-Language-Action (VLA) models against physical sensor attacks, which is crucial to ensuring their secure deployment in the real world. To achieve efficient and large-scale evaluation, we construct a ``Real-Sim-Real'' framework that automatically simulates physics-based sensor attack vectors and validates them on real robotic systems. We also propose an adversarial training defense to mitigate these attacks. We show that existing VLA models are highly vulnerable to physical sensor attacks. Such attacks can severely degrade model performance, resulting in erroneous or hazardous behaviors. Consequently, this vulnerability poses a direct and significant security threat to real-world applications.

\section{Acknowledgments}
We thank the anonymous shepherd and reviewers for their valuable comments. This work is supported by the National Natural Science Foundation of China (NSFC) Grant 62222114.

\bibliography{aaai2026}

\appendix

\section{Appendix A: Sensor Attack Simulation Implementation Details}
We provide the detailed implementation logic for the six camera sensor attacks simulated in our study. For each attack, we present the core algorithm and define its key parameters based on the code provided.

\subsection{Laser Blinding}
This attack simulates the sensor being overwhelmed by a bright laser source. The implementation achieves this by blending the original image with a pre-recorded laser pattern image, effectively simulating a blinding and over-exposure effect.

\begin{algorithm}[H]
\caption{Laser Blinding Attack}
\label{alg:laser_blinding}
\begin{algorithmic}[1]
\REQUIRE $I_{\text{orig}}$: original image array; $I_{\text{laser}}$: laser pattern array; $\alpha$: blending weight
\ENSURE $I_{\text{attacked}}$: attacked image array
\STATE $ I_{\text{attacked}} \gets I_{\text{orig}} \cdot (1 - \alpha) + I_{\text{laser}} \cdot \alpha $
\STATE $ I_{\text{attacked}} \gets \mathrm{clip}\left(I_{\text{attacked}},\, 0,\, 255\right) $
\RETURN $I_{\text{attacked}}$
\end{algorithmic}
\end{algorithm}

\noindent\textbf{Parameters:}
\begin{itemize}
    \item $\alpha$: blending factor in $[0,1]$. Higher values assign more weight to the laser pattern.
\end{itemize}

\subsection{Light Projection}
This attack simulates projecting a malicious pattern onto the scene or directly onto the camera lens. The implementation overlays a specified watermark image with adjustable transparency onto the original image.

\begin{algorithm}[H]
\caption{Light Projection Attack}
\label{alg:light_projection}
\begin{algorithmic}[1]
\REQUIRE $I_{\text{orig}}$: original image array; $I_{\text{wm}}$: watermark pattern; $t$: transparency; $x_{\text{off}}$: horizontal offset; $y_{\text{off}}$: vertical offset
\ENSURE $I_{\text{attacked}}$: attacked image array
\STATE $ pos_x \gets \dfrac{I_{\text{orig}}.\mathrm{width} - I_{\text{wm}}.\mathrm{width}}{2} + x_{\text{off}} $
\STATE $ pos_y \gets \dfrac{I_{\text{orig}}.\mathrm{height} - I_{\text{wm}}.\mathrm{height}}{2} + y_{\text{off}} $
\STATE Place $I_{\text{wm}}$ at $(pos_x, pos_y)$ on $I_{\text{orig}}$ with transparency $t$ to obtain $I_{\text{attacked}}$
\RETURN $I_{\text{attacked}}$
\end{algorithmic}
\end{algorithm}

\noindent\textbf{Parameters:}
\begin{itemize}
    \item $t$: transparency factor in $[0,1]$, controlling the opacity of the projected image (higher values produce a more visible pattern).
    \item $x_{\text{off}}, y_{\text{off}}$: horizontal and vertical offsets (in pixels) from the image center for placing the projected image.
\end{itemize}

\subsection{Laser Color Strip}
This attack exploits the rolling shutter effect of the CMOS sensors to inject colored stripes. The simulation adds color to a horizontal band in the image, with the intensity of the color determined by a 2D Gaussian-like function.

\begin{algorithm}[H]
\caption{Laser Color Strip Attack}
\label{alg:laser_color_strip}
\begin{algorithmic}[1]
\REQUIRE $I_{\text{orig}}$: original image array; $(\rho_R, \rho_G, \rho_B)$: color percentages for red, green, and blue channels; $S$: strength factor
\ENSURE $I_{\text{attacked}}$: attacked image array
\STATE $ h, w \gets \text{height}(I_{\text{orig}}), \text{width}(I_{\text{orig}}) $
\STATE $ row_{\text{start}} \gets \dfrac{h}{3}$, $row_{\text{end}} \gets \dfrac{2h}{3}$
\FOR{$i \gets 0$ \TO $w$}
    \FOR{$j \gets row_{\text{start}}$ \TO $row_{\text{end}}$}
        \STATE $\mathcal{I} \gets \mathcal{I}\big(p = j, q = i, \text{start} = row_{\text{start}}, \text{strength} = S\big)$
        \STATE $ I_{\text{attacked}}[j, i, R] \gets I_{\text{orig}}[j, i, R] + \mathcal{I} \cdot \rho_R $
        \STATE $ I_{\text{attacked}}[j, i, G] \gets I_{\text{orig}}[j, i, G] + \mathcal{I} \cdot \rho_G $
        \STATE $ I_{\text{attacked}}[j, i, B] \gets I_{\text{orig}}[j, i, B] + \mathcal{I} \cdot \rho_B $
        \STATE $ I_{\text{attacked}} \gets \mathrm{clip}\left(I_{\text{attacked}},\, 0,\, 255\right) $
    \ENDFOR
\ENDFOR
\RETURN $I_{\text{attacked}}$
\end{algorithmic}
\end{algorithm}

\noindent\textbf{Intensity Function:} The intensity function $\mathcal{I}(p, q)$ is defined as:

\resizebox{\linewidth}{!}{$
\mathcal{I}(p, q) = S \times \exp\left(-\frac{1}{2(1 - \rho^2)} \left(\frac{(p - c_p)^2}{\sigma_p^2} + \frac{(q - c_q)^2}{\sigma_q^2} - \frac{2 \rho (p - c_p)(q - c_q)}{\sigma_p \sigma_q}\right)\right)
$}

where:
\begin{itemize}
    \item $S$:  strength factor.
    \item $(p, q)$: pixel coordinates.
    \item $(c_p, c_q)$: center coordinates of the effect.
    \item $(\sigma_p, \sigma_q)$: standard deviations related to image height and width.
    \item $\rho$: correlation between the pixel coordinates.
\end{itemize}

\subsection{EM Color Strip}
This attack simulates EMI transmission errors that disrupt color decoding. The implementation applies a fixed, incorrect color transformation to alternating horizontal stripes of the image.

\begin{algorithm}[H]
\caption{EM Color Strip Attack}
\label{alg:em_color_strip}
\begin{algorithmic}[1]
\REQUIRE $I_{\text{orig}}$: original image array; $N_{\text{stripes}}$: number of stripes
\ENSURE $I_{\text{attacked}}$: attacked image array
\STATE $ h,w \gets \text{height}(I_{\text{orig}}),\text{width}(I_{\text{orig}}) $
\STATE $ \text{step} \gets \dfrac{h}{N_{\text{stripes}}} $
\FOR{$x \gets 0$ \TO $w$}
    \FOR{$d \gets 0$ \TO $h - \text{step}$}
        \IF{$\left\lfloor \dfrac{d}{2 \cdot \text{step}} \right\rfloor$ is even}
            \FOR{$y \gets d$ \TO $d + \text{step}$}
                \STATE $(R_{\text{old}}, G_{\text{old}}, B_{\text{old}}) \gets I_{\text{orig}}[y, x]$
                \STATE $ R_{\text{new}} \gets \text{clip}(G_{\text{old}} \times 2.5, 0, 255) $
                \STATE $ G_{\text{new}} \gets \text{clip}\left(\dfrac{R_{\text{old}} + B_{\text{old}}}{2} - 50, 0, 255\right) $
                \STATE $ B_{\text{new}} \gets \text{clip}(G_{\text{old}} \times 2.5, 0, 255) $
                \STATE $ I_{\text{attacked}}[y, x] \gets (R_{\text{new}}, G_{\text{new}}, B_{\text{new}}) $
            \ENDFOR
        \ENDIF
    \ENDFOR
\ENDFOR
\RETURN $I_{\text{attacked}}$
\end{algorithmic}
\end{algorithm}

\noindent\textbf{Parameters:}
\begin{itemize}
    \item $N_{\text{stripes}}$: the total number of horizontal stripes in which to divide the image.
\end{itemize}

\subsection{EM Truncation}
This attack simulates a corrupted image buffer that leads to incorrect frame stitching. The simulation removes a middle portion of the image and appends a section from the end of the original frame, creating a visual discontinuity.

\begin{algorithm}[H]
\caption{EM Truncation Attack}
\label{alg:em_truncation}
\begin{algorithmic}[1]
\REQUIRE $I_{\text{orig}}$: original image array; $r_{\text{trunc}}$: truncation ratio
\ENSURE $I_{\text{attacked}}$: attacked image array
\STATE $ h \gets \text{height}(I_{\text{orig}}) $
\STATE $ \text{line}_{\text{start}} \gets \text{int}(h \times r_{\text{trunc}}) $
\STATE $ \text{line}_{\text{num}} \gets \text{int}(h \times r_{\text{trunc}}) $
\STATE $ \text{line}_{\text{end}} \gets \text{line}_{\text{start}} + \text{line}_{\text{num}} $
\STATE $I_{\text{temp}} \gets \text{delete rows from } \text{line}_{\text{start}} \text{ to } \text{line}_{\text{end}} \text{ in } I_{\text{orig}}$
\STATE $I_{\text{append}} \gets \text{last } \text{line}_{\text{num}} \text{ rows of } I_{\text{orig}}$
\STATE $I_{\text{attacked}} \gets \text{append } I_{\text{append}} \text{ to } I_{\text{temp}}$
\RETURN $I_{\text{attacked}}$
\end{algorithmic}
\end{algorithm}

\noindent\textbf{Parameters:}
\begin{itemize}
    \item $r_{\text{trunc}}$: a float that determines the starting position and the number of lines to be truncated and replaced.
\end{itemize}

\subsection{Ultrasound Blur}
This attack simulates resonance in the camera's IMU, leading to incorrect motion compensation and a blurred image. The simulation applies a complex motion blur, where each output pixel is the average of pixels along a calculated motion path involving rotation, displacement, and scaling.

\begin{algorithm}[H]
\caption{Ultrasound Blur Attack}
\label{alg:ultrasound_blur}
\begin{algorithmic}[1]
\REQUIRE $I_{\text{orig}}$: original image; $\theta$: angle; $(d_x, d_y)$: displacements; $S$: scale
\ENSURE $I_{\text{attacked}}$: attacked image array
\STATE $h,w \gets \text{height}(I_{\text{orig}}),\text{width}(I_{\text{orig}}) $
\STATE $c_x \gets \dfrac{w}{2},\quad c_y \gets \dfrac{h}{2}$
\STATE $\delta \gets \arctan\!\left(\dfrac{d_y}{d_x}\right)$
\STATE $L \gets \sqrt{d_x^2 + d_y^2}$
\STATE $\theta_r \gets \theta \cdot \pi/180$
\FOR{$y \gets 0$ \TO $h$}
  \FOR{$x \gets 0$ \TO $w$}
    \STATE $R \gets \sqrt{(x - c_x)^2 + (y - c_y)^2}$
    \STATE $\alpha \gets \arctan\!\left(\dfrac{y - c_y}{x - c_x}\right)$
    \STATE $X_{\cos} \gets L \cdot \cos \delta - S \cdot R \cdot \cos \alpha$
    \STATE $Y_{\sin} \gets L \cdot \sin \delta - S \cdot R \cdot \sin \alpha$
    \STATE For sample index $n$, compute:
    \STATE \quad $x_{\text{attacked}} = R \cdot \sin(\alpha + n \cdot \theta_r) + n \cdot Y_{\sin} + c_x$
    \STATE \quad $y_{\text{attacked}} = R \cdot \cos(\alpha + n \cdot \theta_r) + n \cdot X_{\cos} + c_y$
    \STATE Accumulate pixel values along the motion path to obtain $I_{\text{attacked}}[y,x]$
  \ENDFOR
\ENDFOR
\STATE $ I_{\text{attacked}} \gets \mathrm{clip}\left(I_{\text{attacked}},\, 0,\, 255\right) $
\RETURN $I_{\text{attacked}}$
\end{algorithmic}
\end{algorithm}

\noindent\textbf{Parameters:}
\begin{itemize}
    \item $\theta, d_x, d_y$: Parameters controlling the linear, rotational, and radial components of the blur.
\end{itemize}

\end{document}